\documentclass{article}
\pdfoutput=1

\usepackage{arxiv}

\usepackage[utf8]{inputenc} 
\usepackage[T1]{fontenc}    
\usepackage{hyperref}       
\usepackage{url}            
\usepackage{booktabs}       
\usepackage{amsfonts}       
\usepackage{nicefrac}       
\usepackage{microtype}      
\usepackage{xcolor}         
\usepackage{amsmath}
\usepackage{graphicx}
\usepackage{url}
\usepackage{paralist}
\usepackage[inline]{enumitem}
\usepackage{enumitem}
\usepackage{graphicx}
\usepackage{subcaption} 
\usepackage{color}

\title{A Phenomenological AI Foundation Model for Physical Signals}

\author{%
\textbf{Jaime Lien} \quad \textbf{Laura I. Galindez Olascoaga} \quad \textbf{Hasan Do\u{g}an} \\
\textbf{Nicholas Gillian} \quad \textbf{Brandon Barbello} \quad
\textbf{Leonardo Giusti} \quad \textbf{Ivan Poupyrev} \\
Archetype AI\\
Palo Alto, CA, USA 94304\\
\texttt{\{jaime, laura, hasan, nick, brandon, leonardo, ivan\}@archetypeai.team}
}

\begin{document}

\maketitle

\begin{abstract}
The objective of this work is to develop an AI foundation model for physical signals that can generalize across diverse phenomena, domains, applications, and sensing apparatuses. We propose a phenomenological approach and framework for creating and validating such AI foundation models. Based on this framework, we developed and trained a model on 0.59 billion samples of cross-modal sensor measurements, ranging from electrical current to fluid flow to optical sensors. Notably, no prior knowledge of physical laws or inductive biases were introduced into the model. Through several real-world experiments, we demonstrate that a single foundation model could effectively encode and predict physical behaviors, such as mechanical motion and thermodynamics, including phenomena not seen in training. The model also scales across physical processes of varying complexity, from tracking the trajectory of a simple spring-mass system to forecasting large electrical grid dynamics. This work highlights the potential of building a unified AI foundation model for diverse physical world processes. 
\end{abstract}

\section{Introduction}
We explore the development of an AI foundation model that can be \textit{universally applied to physical processes of any nature}. Our approach is based on a phenomenological framework, meaning that no prior physical knowledge or inductive bias is introduced. The aim is to construct a single, versatile AI foundation model capable of generalizing across diverse physical phenomena, domains, applications, and sensing apparatuses. This work is inspired by recent advancements in natural language processing (NLP), where generative models based on transformer architectures, such as GPT-4, have demonstrated that a single model trained on a vast corpus of text in self-supervised manner can perform as well as or better than specialized models across a range of tasks \cite{dosovitskiy2021, vaswani2017attention}.

In this paper, we present the design and evaluation of a \textit{physical AI foundation model}, trained on 0.59 billion physical measurements covering a diverse range of real-world processes. The foundation model is paired with lightweight phenomenological decoders that can be individually fine-tuned to generate application-dependent solutions, such as trajectory forecasting. The architecture enables flexible implementation of various use cases while maintaining the foundation nature of the model. Although foundation models have shown significant potential in NLP, image analysis, and general time series data \cite{devlin2018bert, woo2024lotsa, Das2024, Girdhar2023}, their ability to represent and characterize real-world physical processes remains much less explored. Developing such model is one of the contributions of this paper.

The central question we aim to answer 
is whether a foundation model, trained on diverse 
physical data, can characterize physical processes it has not encountered before. We systematically evaluated our model on kinematic and thermodynamic experiments, as well as datasets of complex real-world systems like household electricity use and transformer behavior.
The results were encouraging: the model demonstrated the ability to generalize to physical phenomena and processes it had not seen in training, exhibiting zero-shot inference capabilities. Furthermore, in these use cases, the foundation model performed as well as, or better than, models specifically trained on data from the target physical systems. The experimental evaluation is an additional contribution of this paper.

\begin{figure}
    \centering
    \includegraphics[width=\linewidth]{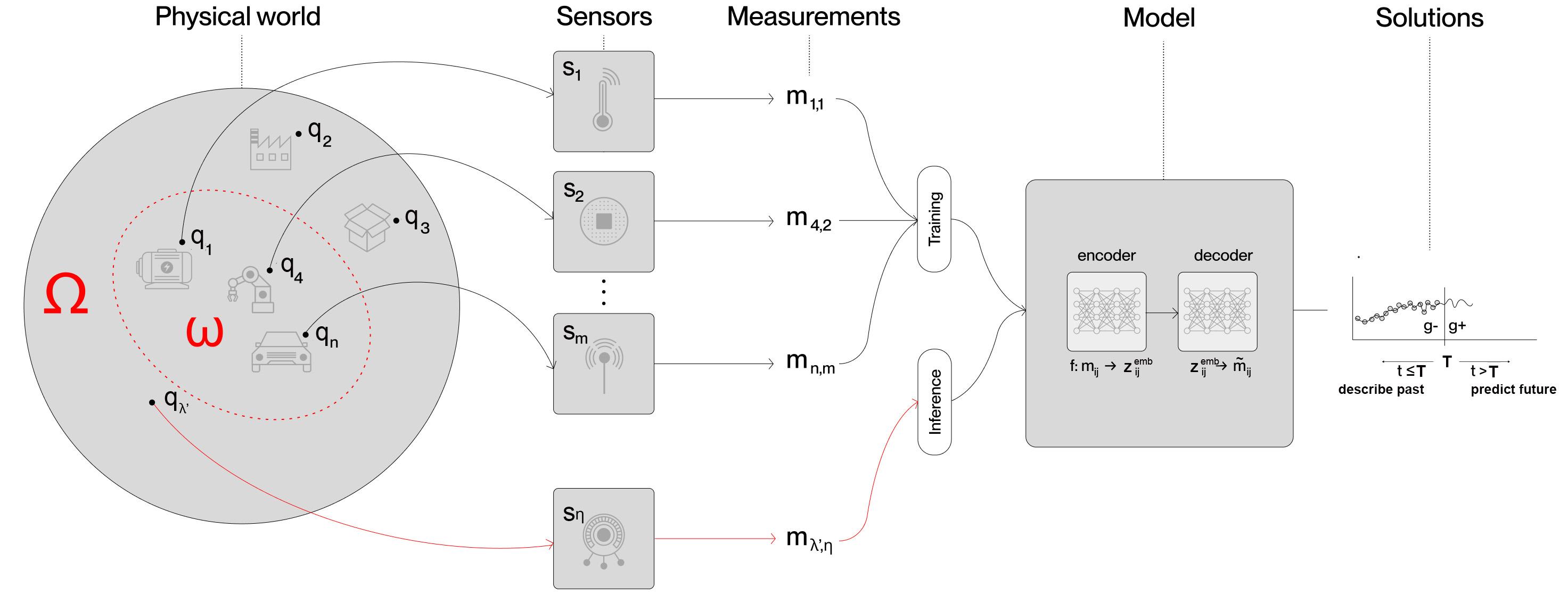}
   \caption{The $\Omega$-Framework describes the relationship between quantities $q_i$, sensors $s_{j}$ and measurements $m_{i,j}$ in the physical world to inform a phenomenological AI model capable of describing past ($t \leq T$) and future ($t > T$) trajectories of any physical quantity of interest $q_i(t)$.}
    \label{fig:fig1-omega}
\end{figure}

\section{Related work}
This work is motivated by the challenges of applying AI methods to interpret real-time physical signals and measurements, a critical requirement for addressing complex, real-world problems. Examples include monitoring and managing large-scale infrastructure systems, such as improving energy efficiency in electrical grids \cite{Akomea-Frimpong2023}, enabling data-driven experimental scientific discoveries \cite{Wang2023}, advancing novel human-computer interaction techniques \cite{Hayashi2021}, and many others. A common approach to designing "physically correct" AI models for these problems involves incorporating \textit{strong inductive biases}, e.g. law of conservation of energy and others \cite{Liu2021, Greydanus2019}. 

While powerful and important, 
this approach creates highly specialized AI models that struggle to generalize to other use cases \cite{Wang2023}.
For example, a fluid motion AI model incorporating the Navier-Stokes equations \cite{Raissi2020} wouldn't work for interpreting radar Doppler images 
\cite{Hayashi2021}. Moreover, many real-world systems cannot be easily formalized by a closed set of equations due to their complexity and the influence of external factors. 
The effectiveness of the strong inductive bias approach in such complex systems remains an open research question.

The \textit{representative} or \textit{phenomenological} \cite{Bas2008} AI models are trained on large datasets of physical measurements without incorporating prior knowledge of physical laws \cite{Iten2020}. These models have proven highly successful in solving complex real-world problems across various application domains and types of sensors, ranging from using accelerometers to analyze human gait \cite{Goncalves-gate} to discovering pulsars from multidimensional radio telescope measurements \cite{Zhu2014}, among many other applications.


However, scalability remains a major challenge for phenomenological AI models. Their effectiveness is inherently limited by the availability of training data, which can be hard to obtain. Additionally, these models are highly sensitive to variations in sensor operational characteristics and pre-processing methods, often requiring further data collection and retraining. Consequently, they are typically narrowly designed and can not easily generalize across different use cases, sensors, or systems. 


In this paper, we explore a foundation model approach, where a model trained on vast amounts of diverse physical data across application domains, sensor types, and physical processes can develop a general representation of physical behaviors. This allows the model to be deployed in use cases with no additional training (i.e., zero-shot) or with a small amount of additional data (i.e., fine tuning). The remainder of this paper details the framework used to formalize the requirements for such a model, the design and the results of our experimental evaluations of the model.

\section{$\boldsymbol{\Omega}$-Framework for building a universal phenomenological model}
\subsection{Physical quantities}
Consider an electric motor whose internal construction and operation
are unknown. Its current state and behavior can be described by a set of observable, time-varying physical quantities, such as temperature, electrical current, rotational speed, torque, etc. Each of these quantities can be represented as $q_i(t)$, where $t$ denotes time. Consequently, any physical process $\lambda$, including our motor, can be represented as the union of all its observable physical quantities, $Q_\lambda = \cup_i q_i(t)$. It is reasonable to postulate that the physical world as a whole can be represented as a superset $\Omega$ that encompasses all possible physical quantities for all objects varying across space and time:
\[
\Omega = \bigcup_{\lambda} Q_\lambda
\]
One could aspire in theory to train a universal phenomenological model on the entirety of $\Omega$. However, this is not practically feasible; even if sufficient computational resources existed to train such a model, there is not a complete dataset representing all physical processes across the physical world. 

Instead, we ask if a compact representative model trained on a subset of observable measurements $ \omega \in \Omega$ can infer and accurately describe the behavior of physical quantity $q_{\lambda'} \notin \omega$ that was not included in the training data set (see Figure \ref{fig:fig1-omega}). 

Our framework considers two important issues in this regard: the observation of latent physical quantities via sensors, discussed in Section \ref{sec:sensors}, and the formulation of the phenomenological model as an AI architecture, described in Section \ref{sec:formulation}.

\subsection{Sensors and physical measurements}
\label{sec:sensors}
The challenge with building models for the physical world is that neither the physical quantities nor their governing laws are inherently known or can be sampled directly. Instead, they can only be \textit{indirectly observed through sensors} $s_j$, which transduce the true physical quantity $q_i(t)$ into electrical signals and corresponding measurements $m_{i,j}(t)$, that is $s_j:q_i(t)\mapsto m_{i,j}(t)$. Importantly, the sensors introduce noise and distortions due to their specific operational characteristics.

The relationship between the observed measurements $m_{i,j}$ and the latent quantity $q_i$ can be characterized by the conditional probability density function $p_{M|Q}(m_{i,j}|q_i)$ describing the distribution of the measurement values given the unknown true quantity values \cite{bertsekas2008introduction}.

The measurements $m_{i,j}(t)$ that are used to train the model are thus incomplete, stochastic observations of the underlying continuum of true physical quantities $q_i(t)$ and can be characterized as: 
\[
p_{M}(m_{i,j}) = \int_{q_i} p_{M|Q}(m_{i,j}|q_i) p_Q(q_i) dq_i, 
\]
 where $p_Q(q_i)$ is the true underlying distribution of $q_i$, is unknown and cannot be directly sampled.

\subsection{Model formulation} 
\label{sec:formulation}

We ask whether it is possible to build a single phenomenological AI model for all physical quantities $q_i \in \Omega$ — that is, a compressed representation of $\Omega$ capable of both \textit{describing past trajectories} and \textit{predicting future trajectories} of any physical quantity of interest $q_i(t)$ as measured by sensor $s_j$. 

This definition is consistent with previous approaches on designing AI models for physical world \cite{Liu2021, Greydanus2019}. This is also analogous to how analytic physical models, such as a system of differential equations governing a pendulum, represent past trajectory and can also be used to predict future behavior.

We therefore formulate the model as a mapping $f$ that encodes the measurements $m_{i,j}(t)$  up to time $T$ into latent representation $z_{i,j}^{\text{emb}}$, which can be accurately decoded into the past and future measured trajectories of $q_i(t)$, denoted $\Tilde{m}_{i,j}(t)$, using decoders $g_-$ and $g_+$ respectively (see Figure \ref{fig:fig1-omega}):
\begin{equation}
\begin{split}
f:\:&m_{i,j}(t) \rightarrow z_{i,j}^{\text{emb}}, \\
g_-:\:&z_{i,j}^{\text{emb}} \rightarrow \Tilde{m}_{i,j}(t),\text{for }t \leq T, \\
g_+:\:&z_{i,j}^{\text{emb}} \rightarrow \Tilde{m}_{i,j}(t),\text{for }t > T,
           \end{split}
\label{eq:mappins}
\end{equation}
such that the mappings $f$, $g_{-}$, and $g_{+}$ hold for all $q_i \in \Omega$ and sensors $p_{M|Q}(m_{i,j}|q_i)$.

The open question that we explore in the rest of the paper is whether these mappings can be learned from a diverse but sparse subset $\omega \in \Omega$ and how well they hold over the much broader space of physical quantities $q_\lambda(t)\notin\omega$, i.e. quantities that were not part of the training set. In other words, if we train the model on a large and diverse set of physical data, will the model be able to generalize to measured quantities that are significantly different from the training set, such as:

\textit{Physical phenomena} that the model was not trained on, for instance, can the model infer or understand thermoelectric behavior even if the training set contains no data related to this phenomena?

\textit{Highly complex physical processes} that cannot be represented in a closed form, for example, can the model accurately predict city-scale meteorological dynamics?

\textit{Physical quantities captured by diverse sensors}, where the physical quantity of interest remains the same, but the measurements differ due to the use of different sensor types, e.g. can the model interpret observations from liquid-based motion sensors if it only saw piezoelectric MEMS sensors in training?


In the rest of the paper, we discuss the model's design and experimental results toward these questions.

\section{$\boldsymbol{\Omega}$-Model design and training}
\begin{figure}
    \centering
    \includegraphics[width=\linewidth]{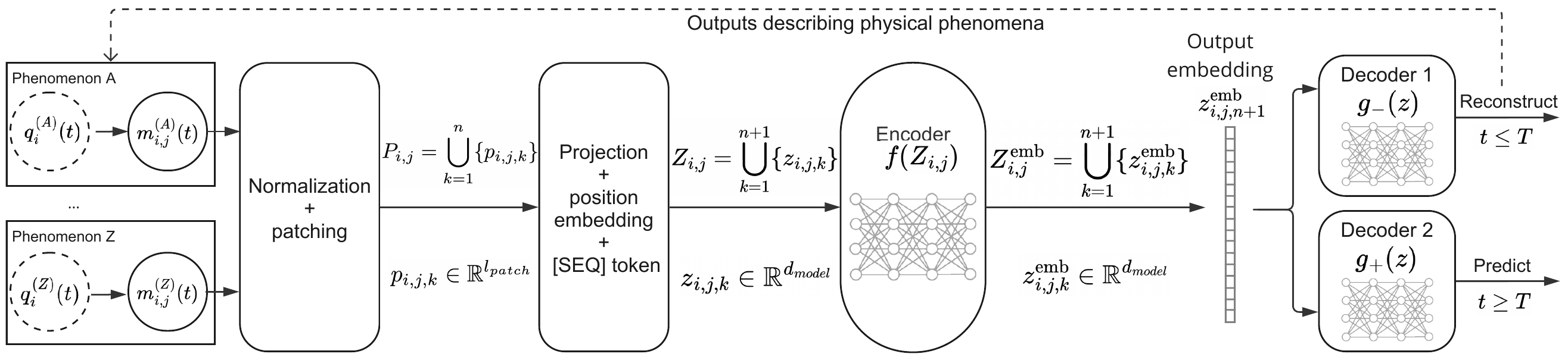}
    \caption{Architecture: Sensor data is normalized and segmented into fixed-length, one-dimensional patches that are used to train a transformer-based encoder network. 
    Compact downstream decoders are then employed to reconstruct and predict the trajectories of the physical quantities of interest.}
    \label{fig:model_high_level}
\end{figure}

This section presents the design and training of proposed model. 
As illustrated in Figure \ref{fig:model_high_level}: sensor measurements are encoded as temporal sequences of one-dimensional patches of fixed length, which are projected into a unified embedding space using a standard transformer architecture \cite{vaswani2017attention, dosovitskiy2021}. The transformer architecture preserves the deep temporal structure of individual patches, and as the encoder is trained on large, diverse sets of sensor data, it effectively learns the underlying temporal patterns of a wide range of complex physical processes, regardless of the specific sensor modality. 

The resulting compact, universal representation can then be employed with relatively lightweight phenomological decoders, resulting in \textit{trajectory prediction} or \textit{forecasting} and \textit{reconstruction} solutions (Figure \ref{fig:model_high_level}). These decoders correspond to the $g_-$ and $g_+$ mapping functions defined in Equation \ref{eq:mappins}.

\subsection{Encoder for physical quantities}

Encoder projects sensor data into universal latent representation by training the transformer-based network on wide variety of diverse physical data. 

\textbf{Patching.} The measurements of physical quantity $q_i(t)$ are captured using sensor $s_j$ during time period $T_0 < t \leq T$, normalized to the range [0,1], and represented as $m_{i,j}(t)$. Each measurement $m_{i,j}(t)$ is then segmented into a sequence of $n$ patches: $P_{i,j} = \bigcup_{k=1}^{n} \{ p_{i,j,k} \}, \text{where patch }p_{i,j,k} \in \mathbb{R}^{l_{patch}}$.
Following related work that employed fixed patch sizes for images \cite{dosovitskiy2021}, we do not adjust the resolution of the patches to match the sampling frequency of the measurements.
This approach increases the amount of available data and simplifies model training. 

\textbf{Projection to the embedding space.} The set of patches $P_{i,j}$ is projected into the embedding space of dimension $d_{model}$ using a linear layer, resulting in a sequence of $n$ patch embeddings $T_{i,j} = \bigcup_{k=1}^n \{t_{i,j,k}\}$. A learnable positional embedding $r_k \in \mathbb{R}^{d_{model}}$ is added to each patch embedding $t_{i,j,k}$ to capture the temporal order of the patches \cite{devlin2018bert}. Each resulting embedding is denoted $z_{i,j,k} = t_{i,j,k}+r_k$ for $k=1,\ldots,n$. Finally, we append one learnable \texttt{[SEQ]} token \footnote{Represents the full input sequence in a single embedding, analogous to the\texttt{[CLS]} token in \cite{devlin2018bert,dosovitskiy2021}.} to the patch embeddings, denoted by $z_{i,j,(n+1)}$. The resulting embedding sequence $ Z_{i,j} = \bigcup_{k=1}^{n+1} \{ z_{i,j,k} \}$ is then input to the transformer encoder.

\textbf{Encoder architecture.} We use a vanilla transformer \cite{vaswani2017attention} with 6 layers of alternating causal multi-head attention heads and feedforward layers. The output of the attention is projected by a learnable linear layer, batch normalized and summed with residual connections; and  the feedforward network is an MLP. The output of the transformer is a sequence of embeddings $Z_{i,j}^{\text{emb}} = \bigcup_{k=1}^{n+1} \{z_{i,j,k}^{\text{emb}}\}$ from which we extract the $(n+1)$th one, corresponding to the location of the appended \texttt{[SEQ]} token. This extracted embedding, denoted $z_{i,j}^{\text{emb}}$ for simplicity, constitutes a phenomenological representation of the physical quantity $q_i(t)$ as measured by sensor $s_j$. 

\subsection{Phenomenological decoders }

To apply and validate the phenomenological representation, the embedding vector $z_{i,j}^{\text{emb}}$ is input into decoders $g_{-}$ and $g_{+}$. Decoder $g_{-}$ is a learned mapping from the phenomenological representation 
$z_{i,j}^{\text{emb}}$ to a reconstruction of past measurements, $\Tilde{m}_{i,j}(t)$ for $t \leq T$.
Decoder $g_{+}$ is a learned mapping from 
$z_{i,j}^{\text{emb}}$ to a prediction of future measurements, $\Tilde{m}_{i,j}(t)$ for $t > T$.

We use lightweight two-layer MLP architectures for both decoders. A similar architecture can be used for other tasks such as classification. The motivating idea is that the encoder should capture and represent the salient information in the physical measurement, allowing the downstream decoders to have a relatively small number of parameters and making them easily trainable. For a detailed description of the decoders' architecture, see Appendix \ref{sec:appendix_decoders}. 

\subsection{Training}

Our training set $\omega$ consists of physical  measurements from 41 publicly available datasets covering a wide variety of phenomena, quantities, and sensors, including river flow, solar power generation, precipitation, solar activity, and more (see Appendix \ref{sec:appendix_training} for a full list). In total, our training set comprises 595,730,272 observations of the physical world. We pretrain the model using two unsupervised tasks: forecasting and auto-encoding. The pretraining tasks are performed concurrently by training the encoder end-to-end with decoder $g_{-}$ and decoder $g_{+}$ in a dual-head configuration. The mean square error (MSE) of the forecast and the reconstruction are weighted 0.6 and 0.4 respectively in the pretraining loss function.
We use a batch size of 1024, learning rate of 1e-3, with an AdamW optimizer. On a single Nvidia A100 GPU, model pre-training takes 6 hours.

We fine-tune the model by freezing the pre-trained encoder weights and training just the applicable decoder on the target dataset.

\section{Experimental validation}


This section presents two sets of experiments 
evaluating the generalizability of the model and its ability to accurately describe the behaviors of real-world physical systems $Q_{\lambda'}\in \Omega$ that the model \textit{did not observe during training}, i.e. $Q_{\lambda'} \notin \omega$ (see Figure \ref{fig:fig1-omega}).

For the first set of experiments (Section \ref{sec: canonical_exp_description}), we selected two \textit{canonical systems} -- a mechanical oscillator and a thermodynamic system -- chosen for their simplicity in interpreting results, as well as the use of such experiments as standard benchmarks (e.g., see \cite{Cranmer2020, Liu2021}). In the second set of experiments (Section \ref{sec: complex_exp_description}), we evaluate whether the model can generalize to complex non-analytic, \textit{real-world systems}, 
and for this purpose, we utilized publicly available datasets. 

We considered two downstream tasks for evaluation: (1) \textit{trajectory reconstruction}, which assesses the model's ability to project the embedding space produced by the encoder onto a trajectory describing the past behavior of the system (corresponding to decoder $g_-$ in Figures \ref{fig:fig1-omega} and \ref{fig:model_high_level}); and (2) \textit{future trajectory prediction}, which evaluates the model's ability to forecast the future behavior of the system and corresponds to decoder $g_+$ in Figures \ref{fig:fig1-omega} and \ref{fig:model_high_level}. In particular, we evaluated the complex physical process experiments on both task (1) and (2), while  we only evaluated the canonical experiments on task (2), and leave the evaluation of these experiments on task (1) for future work.


\subsection{Canonical systems experiments} 
\label{sec: canonical_exp_description}
\begin{figure}
    \centering
    \includegraphics[width=0.9\linewidth]{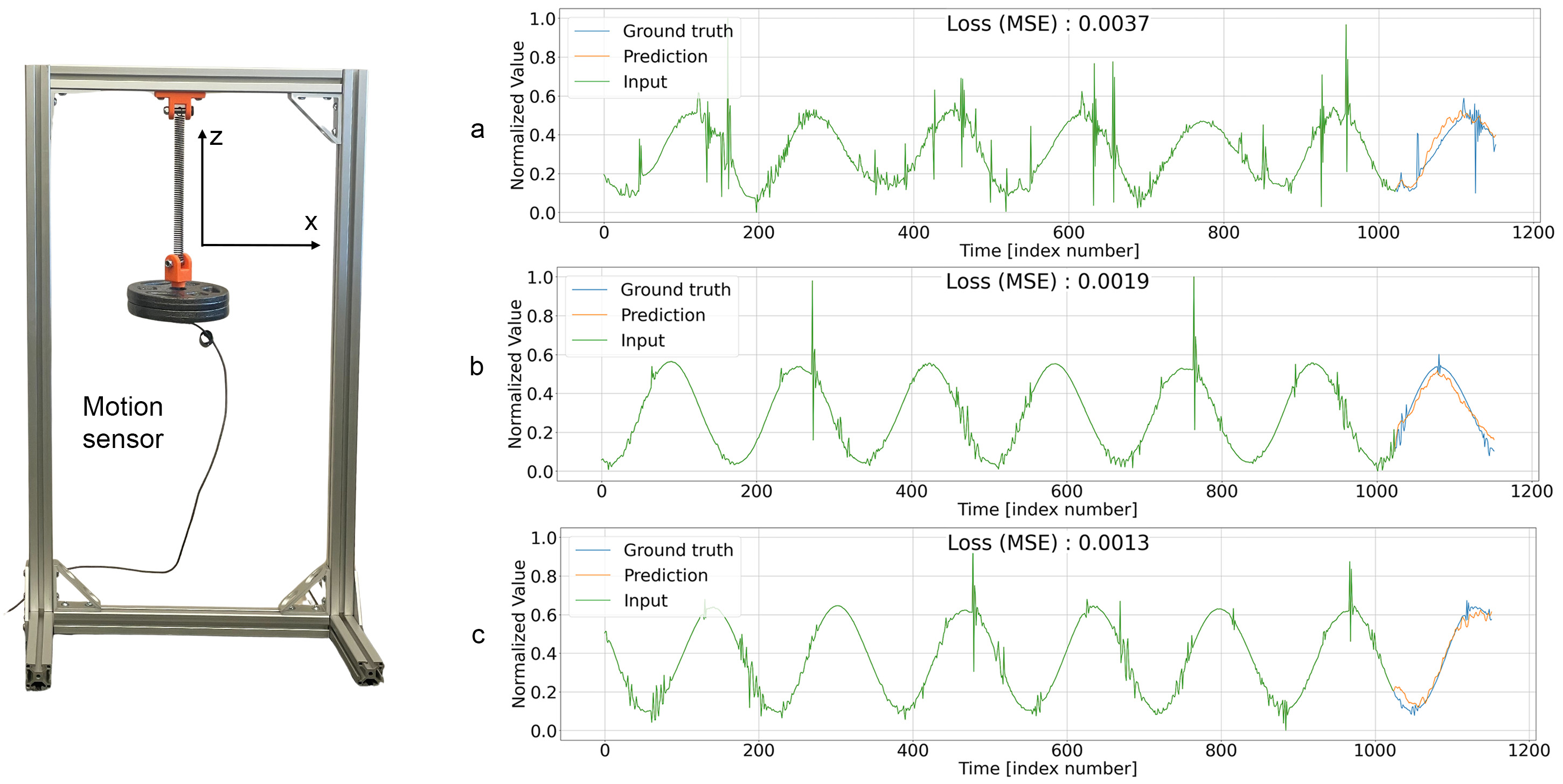}
    \caption{Harmonic oscillator experiment. Left: experimental set up consist of a spring-mass system. Right: Sample forecasts for three regions of oscillatory behavior: (a) semi-chaotic dampened oscillation, (b) transition to dampened harmonic oscillation, and (c) dampened harmonic oscillation.}
    \label{fig:mecha_experiment}
\end{figure}

    \subsubsection {Experimental design}
All canonical experiments were conducted by collecting data using experimental physical apparatuses described below, subsequently evaluated using a single Nvidia A100 GPU.


\textbf{Harmonic oscillator.}  We implemented a mechanical oscillator through a hanging spring-mass apparatus compising a 0.65-inch long steel spring with a spring rate of 1.9 lbs/inch and a 10-pound mass attached to the end of the spring (Figure \ref{fig:mecha_experiment}). When manually perturbed by pulling the mass in the negative z- and y-directions by 2 inches and releasing it, this system initially exhibits chaotic behavior (as an elastic pendulum), settling later into classic dampened harmonic oscillator. The resulting oscillations were recorded using an STMicro LSM6DS3 accelerometer, attached to the bottom of the mass, sampling z-axis acceleration at 208 Hz with 16-bit fixed-point resolution. In total, 19360 acceleration measurements were collected, correspond to 93 seconds of recording. 
To evaluate the model's zero-shot forecasting, we segmented the recording into 156 windows of 1024 samples (4.92 seconds), with a sliding stride of 128 samples. For each segment, we used the 1024 samples as the input to the model and computed the forecast of the subsequent 128 points.

\begin{figure}
    \centering
    \includegraphics[width=0.9\linewidth]{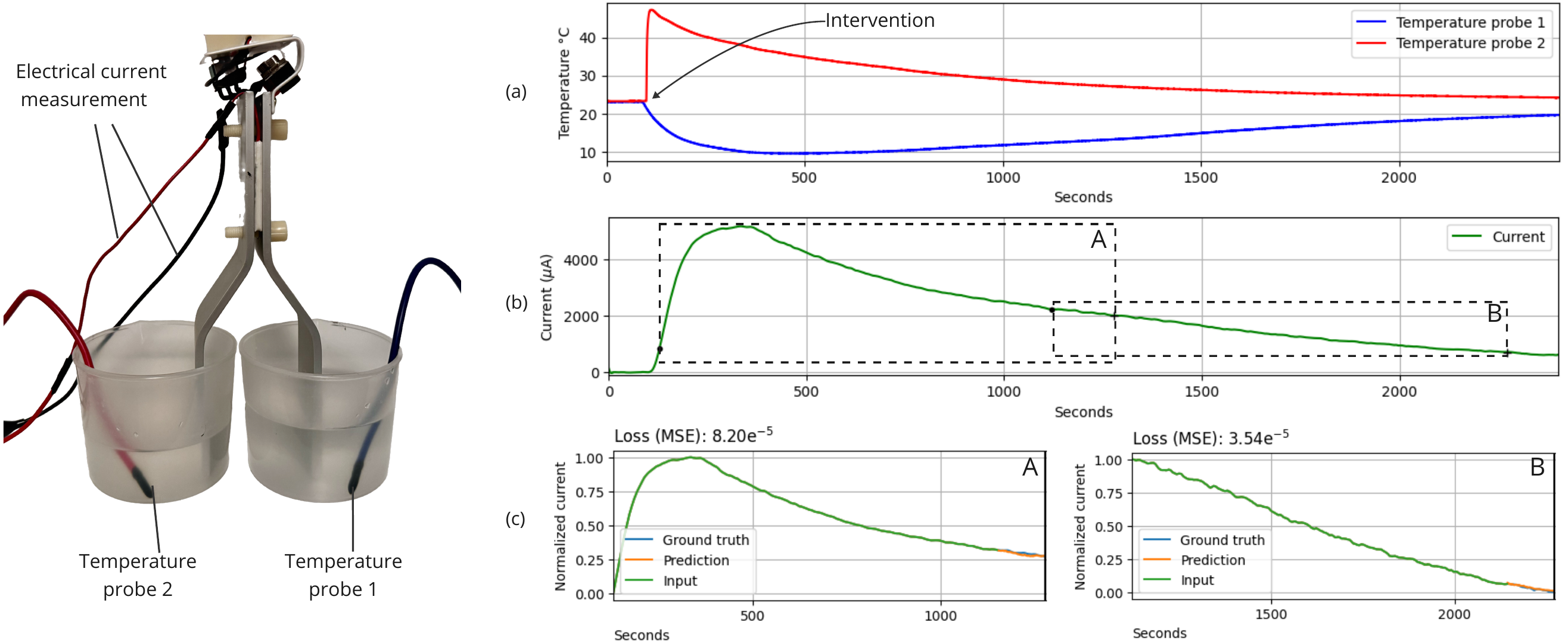}
    \caption{Thermodynamic experiment. Left: experimental set up; 
    Right: example of (a) temperature of the two water baths, (b) the output electrical current induced by the temperature differential, (c) example forecasts over two windows labeled A and B, respectively.}
    \label{fig:thermo_experiment}
\end{figure}

\textbf{Thermoeletric demonstrator.} The behavior studied in this experiment is based on the Seebeck effect: 
 a pair of aluminum strips submerged in water baths at different temperatures induce a temperature gradient (Figure \ref{fig:thermo_experiment}). When applied to a semiconductor block (Bismuth Telluride), this gradient generates an electrical potential, which we sample to evaluate our model through 40 minutes-long experiments. 
An INA219DC Current Monitor was used to measure the output current measured in $\mu$A, at a rate of 10 Hz with 12-bit fixed-point resolution, 
resulting in 24000 samples, corresponding to 120 minutes total. To account for the slow evolution of the thermoelectric effect and the high-frequency noise from the sensor, we downsampled the measurements to 1 Hz and applied a one-dimensional smoothing filter. For the zero-shot forecasting task, we used a context window of 1,024 frames ($\sim$17 minutes) predicted over 128 frames ($\sim$2.1 minutes). To evaluate different context windows, we segmented each recorded signal by sliding a window every 16 seconds.


\subsubsection {Results of canonical experiments}
In this section, we report the model's forecasting performance on the collected  data. The model was not fine-tuned on this data, and the same weights were used across all analyses. The mean squared error (MSE) is reported to estimate the discrepancy between the model output and the ground truth.

\textbf{Harmonic oscillator experiments.} Figure \ref{fig:mecha_experiment} shows examples of the model's forecast in three regions of oscillatory behavior: (a) semi-chaotic dampened oscillation, (b) transition to dampened harmonic oscillation, and (c) dampened harmonic oscillation. The MSE of the forecasted z-axis acceleration in these regions is 0.0037, 0.0019, and 0.0013 respectively. In general, the forecasting performance improves as the kinematic motion transitions from chaotic elastic pendulum motion to more regular dampened harmonic motion.  The average MSE across all 156 windows of the recording was 0.00968. We found that windows with lower signal-to-noise ratio, e.g. where the amplitude of oscillation was smaller, tended to be the largest contributors to the overall error.

\textbf{Thermodynamic experiments.} Figure \ref{fig:thermo_experiment} shows two examples of forecasting system behavior over different windows from one of the recordings discussed earlier. They demonstrate that our pre-trained model is capable of accurately performing zero-shot forecasts on physical phenomena that it was not trained on, achieving MSE values of $8.2\mathrm{e}^{-5}$ and $3.5\mathrm{e}^{-5}$, respectively. Overall, the average MSE across the 78 windows of each recording were $15\mathrm{e}^{-5}$, $34\mathrm{e}^{-5}$, and $37\mathrm{e}^{-5}$, respectively. While the MSE in most windows was smaller than average, there are corner cases where the prediction fails to capture the correct temporal pattern (see Appendix \ref{sec:thermo_corner}). This is likely due to the encoder over-fitting on periodic patterns in the training data, as well as residual high-frequency noise from the sensor.

Optimizing the model to address these 
corner cases is beyond the scope of this paper and can be done by expanding the dataset or fine-tuning a task-specific decoder. We leave this for future research.

\subsection{Complex physical processes} 
\label{sec: complex_exp_description}

\subsubsection {Experimental definition}
We considered measurements of five real-world physical processes that were not included in the training dataset, drawn from different domains and characterized with various sensors: 
\vspace{-2mm}
\begin{enumerate}[label=\alph*\upshape)]
\item Daily minimum temperature in Melbourne, Australia, from 1981 to 1990 \cite{dailymintemp}. \item Hourly country-wide power consumption of Turkey, from 2015 to 2020, in MWh \cite{turkey}. \item Hourly electricity consumption of 321 Portuguese clients from 2012 to 2014, in kW \cite{electricity}.\item Oil temperature of an electrical transformer from a province of China from July 2016 to July 2018, measured in fifteen minute intervals \cite{ett}. \item Meteorological parameters, such as air temperature, humidity and wind speed, 
measured 
in 
Germany, every 10 minutes over the year of 2020 \cite{weather}.
\end{enumerate}

For each experiment, we evaluated the $\Omega$-model’s zero-shot reconstruction and forecasting performance in terms of MSE. We also evaluated the improvement in performance resulting from fine-tuning the $g_{-}$ and $g_{+}$ decoders using data from the target domain. Finally, as a comparison, we trained the network solely on the target dataset and again computed the reconstruction and forecasting performance. For the fine-tuned and target-trained models, we follow the standard process of isolating the most recent 10\% of the target measurements for evaluation, the prior 10\% of target measurements for validation, and the earliest 80\% of target measurements for training. This 
scheme allows us to evaluate model handling of a potential shift in the underlying distribution $p_Q(q_i)$ over time.

\subsubsection {Results}

\begin{figure}
   \centering
   \includegraphics[width=\linewidth]{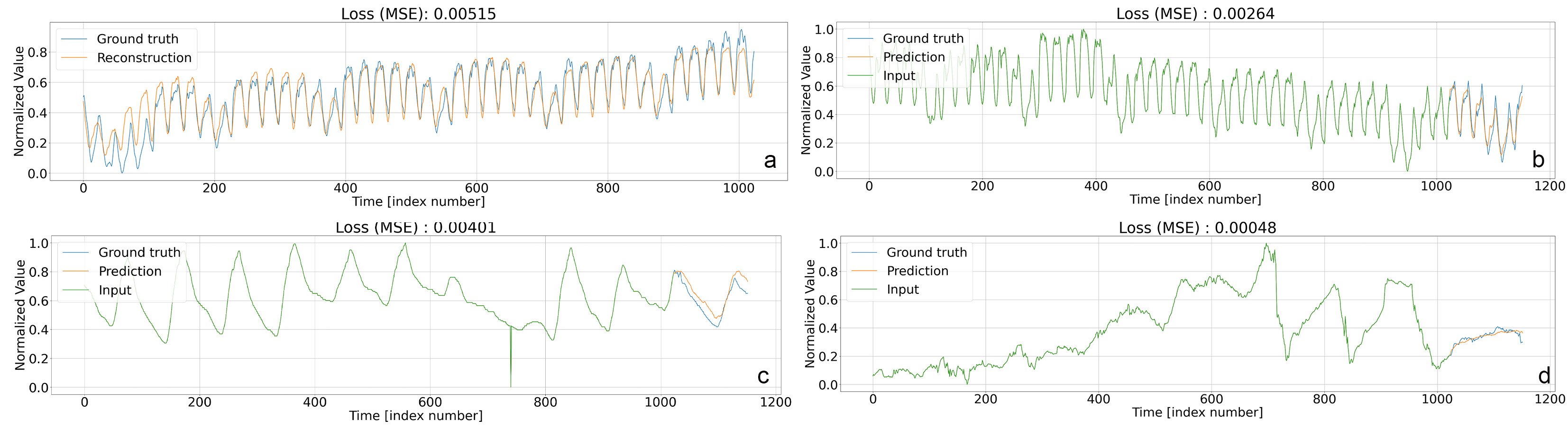}
  \caption{Sample graphs for complex real-world processes. a) and b) Reconstruction and forecasting of Turkey's power consumption; c) Forecasting the oil temperature of an electrical transformer; d) Forecasting water vapor concentration in Germany.}
   \label{fig:complex_graphs}
\end{figure}

Figure \ref{fig:complex_graphs} shows examples of zero-shot forecasting and reconstruction for real-world measurements of physical processes. These examples demonstrate the model's ability to phenomenologically describe and predict diverse physical signals that it was not explicitly trained on and that vary in sampling rate, spectral content, and noise characteristics, all with a single set of weights.

\begin{figure}
   \centering
   \includegraphics[width=\linewidth]{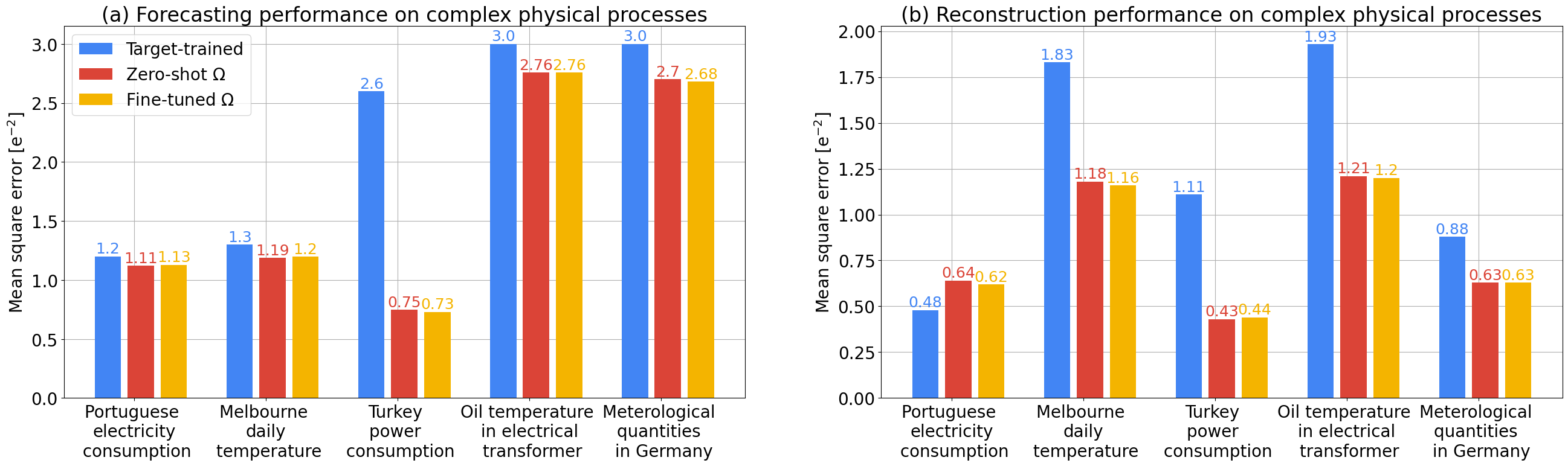}
  \caption{Comparison of (a) forecasting and (b) reconstruction mean square error for the zero-shot $\Omega$-model, fine-tuned $\Omega$-model, and target-trained models.}
   \label{fig:complex_barchart}
\end{figure}

Figure \ref{fig:complex_barchart} (a) compares the MSE when predicting future measurements from the phenomenological representation. Here, the zero-shot $\Omega$-model outperforms the target-trained model in each experiment, indicating that the trajectory of these physical processes can be learned solely from other phenomena. Across all the experiments, the zero-shot $\Omega$-model's MSE was on average 23\% less than the MSE for the target-trained model. We observed a further improvement of 0.8\% when fine-tuning the $\Omega$ $g_{+}$ decoder with target data.

Figure \ref{fig:complex_barchart} (b) shows the MSE when reconstructing physical measurements from the phenomenological representation. In most experiments, the zero-shot $\Omega$-model significantly outperforms the target-trained model, with the average zero-shot MSE 34\% better than the target-trained models. Fine-tuning the $\Omega$ $g_{-}$ decoder on the target data results in a further improvement of 2\%.

We hypothesize that the $\Omega$-model outperforms target-trained models due to its ability to handle shifts in the underlying distribution of physical quantities $p_Q(q_i)$ over time, which is a common occurrence in complex, real-world processes. The target-trained models are less able to adapt to such distribution shifts and more prone to overfitting, as their training dataset samples from a much narrower distribution of measurements compared to the $\Omega$-model.

The relatively small performance difference between the zero-shot $\Omega$-model and the fine-tuned $\Omega$-model warrants further investigation. 
The limited amount of target data available for fine-tuning may be a contributing factor. A deeper study of fine-tuning approaches is left for future work.


\section{Conclusions and Discussion}
\textbf{Unified AI model across physical phenomena}:
A single AI foundation model, pretrained on diverse physical signal data, can generalize across different physical phenomena, sensor types, and complex application domains where the governing mechanisms and underlying probability distributions are unknown. Although preliminary, these findings suggest the possibility of a \textit{single AI model that can handle different areas of physics}. 

\textbf{Zero-shot AI capabilities for physical phenomena}:
The model demonstrated the ability to encode, represent, and forecast classes of physical behaviors it had not encountered during training, i.e. \textit{zero-shot learning of physical phenomena}. For example, the model was able to predict thermoelectric dynamics despite not being trained on thermoelectric effects, or performing better on zero-shot prediction of transformer oil heating than models trained specifically on this target data. 

\textbf{Parallels with Large Language Models (LLMs)}:
Our model showed similarities to LLMs, which can zero-shot match or surpass specialized models, but in the domain of physical signals. It has been argued that LLMs trained on large text corpora capture the semantic meaning and relationships in language, beyond just grammar and syntax patterns. Could there be a similar possibility in the physical world, where an AI model can capture the underlying structure of physical processes? 

\subsection{Practical implications}


\textbf{Out-of-the-box sensor intelligence}: The pre-trained foundation model would be able to analyze sensor measurements even with limited training data, making it useful in domains where data collection is challenging. This could give rise to sensor-agnostic systems that can work with specialized sensors for which data collection is difficult.

\textbf{Adaptation through decoders and fine-tuning}: The architecture we proposed involves designing and training a foundation encoder with lightweight, fine-tuned decoders for specific phenomenological solutions.
This approach allows the model to be easily adapted to a range of applications without requiring significant modifications to its core structure or large amounts of data for fine-tuning.

\textbf{Autonomous learning from observations}: The self-supervised training of our model suggests its ability to learn physical behaviors directly from observational data. This capability opens the door to developing autonomous systems that can independently adapt to new operational environments or requirements without explicit human data labeling, simplifying the deployment of AI solutions.

\textbf{Advances in scientific discovery}: We are particularly excited about the potential of phenomenological physical AI models in scientific discovery and experimental analysis, especially where data acquisition is limited. Such models could dramatically 
accelerate experimental research.


\subsection{Future work}
This work represents an early exploration of unified AI models for physical signals with many open questions and limitations that warrant future work. While sensor agnosticism was introduced into our framework, this important aspect remains underexplored. Sensor characteristics can introduce significant attenuation and distortion of the true signal, affecting the accuracy of physical measurements. We also observed that measurement sampling frequency had a notable impact on the model’s performance, indicating the need for better handling of varying rates. Future work should examine model performance on non-periodic signals, edge cases, and anomalies, as well as the effects of human intervention. Exploring alternative designs for downstream decoders is another promising area, as it may lead to numerous practical applications.




We hope that this work will stimulate discussion about the applicability of generative AI foundation models in the physical world and encourage further exploration by the community.

\bibliography{mybib}{}
\bibliographystyle{abbrv}

\appendix
\section{Appendix / supplemental material} 
\label{sec:appendix}

\subsection{Phenomenological decoders}
\label{sec:appendix_decoders}

The embedding vector $\mathbf{Z}^{\text{emb}}_{i, n+1}$ is then input into task-specific decoders. In this work, we examine two exemplar tasks: reconstruction and forecasting. Both decoders are lightweight MLPs with the same architecture, except for the dimensionality of the intermediate and last layers. A similar architecture can be used for other tasks such as classification. The motivating idea is that the encoder should capture and represent the salient information in the sensor measurement, allowing the downstream task-specific decoders to have a relatively small number of parameters and making them easily trainable. The decoder MLP architecture consists of a layer normalization step, a learnable linear layer $\mathbf{W}_{d,1} \in \mathbb{R}^{d_\text{model} \times d_\text{int}}$, ReLU activation, and a second learnable linear layer $\mathbf{W}_{d,2} \in \mathbb{R}^{d_\text{int} \times d_\text{out}}$. The intermediate dimension $d_\text{int}$ and output dimension $d_\text{out}$ vary according to the task. For reconstruction, $d_\text{out}$ is equal to the number of points in the input measurement, $d_\text{out} = nl_\text{patch}$ and $d_\text{int} = \frac{d_\text{out}}{2}$. For forecasting, $d_\text{int} = d_\text{model}$ and $d_\text{out} = l_\text{pred}$ where $l_\text{pred}$ denotes the desired length of the prediction window.

\subsection{Thermoelectric experiments: examples of corner cases}
\label{sec:thermo_corner}

\begin{figure}[ht]
    \centering
    \begin{subfigure}[b]{0.5\textwidth}
        \centering
        \includegraphics[width=\textwidth]{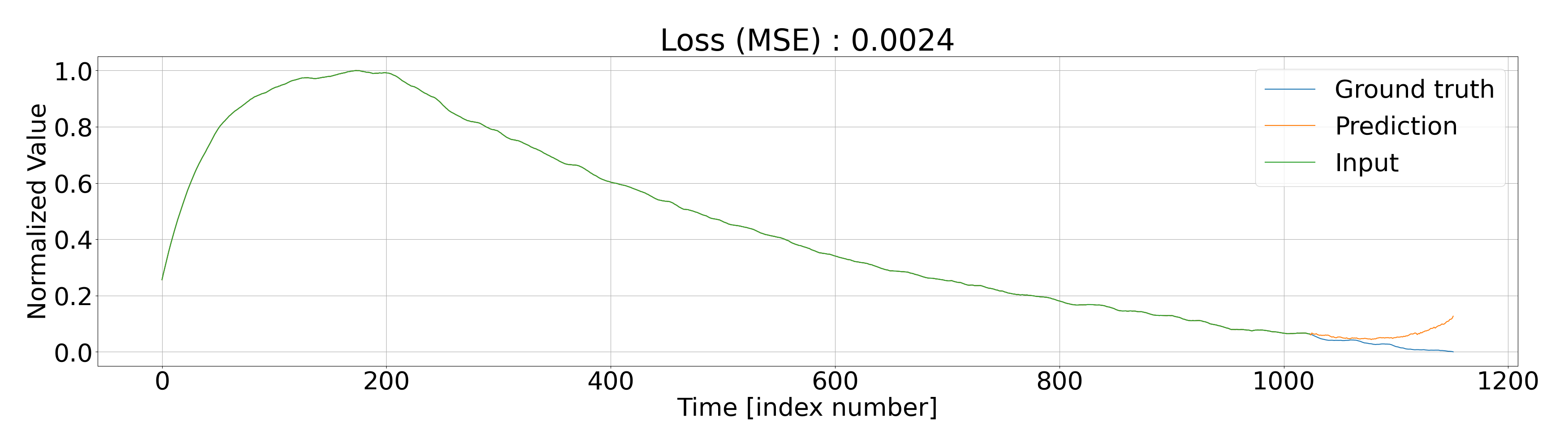}
        \caption{Corner case 1}
        \label{fig:subfig1}
    \end{subfigure}
    \hfill
    \begin{subfigure}[b]{0.45\textwidth}
        \centering
        \includegraphics[width=\textwidth]{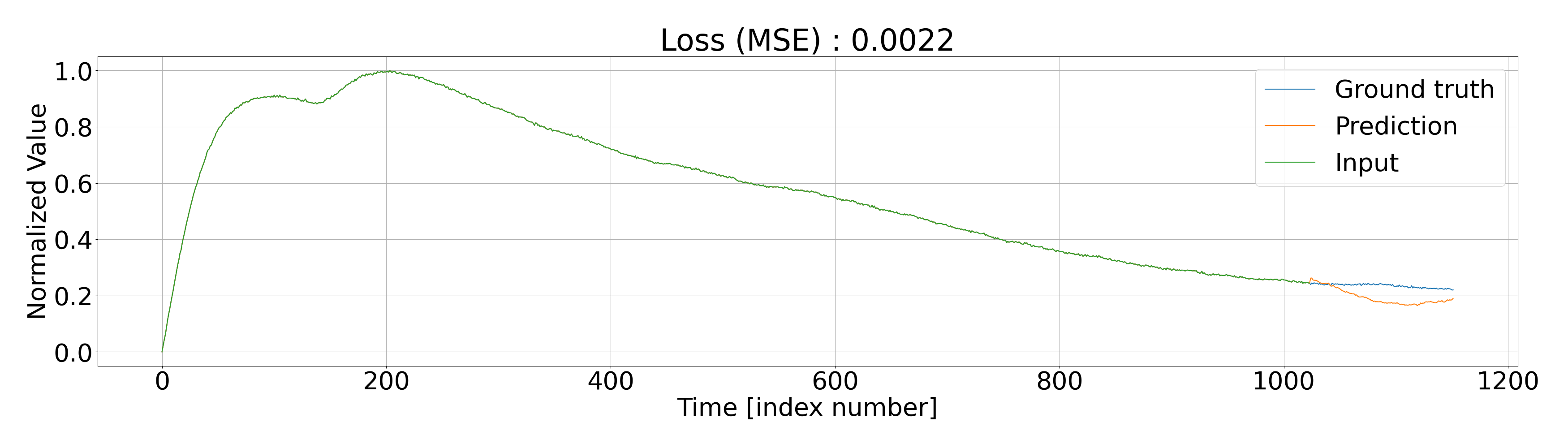}
        \caption{Corner case 2}
        \label{fig:subfig2}
    \end{subfigure}
        \hfill
    \begin{subfigure}[b]{0.5\textwidth}
        \centering
        \includegraphics[width=\textwidth]{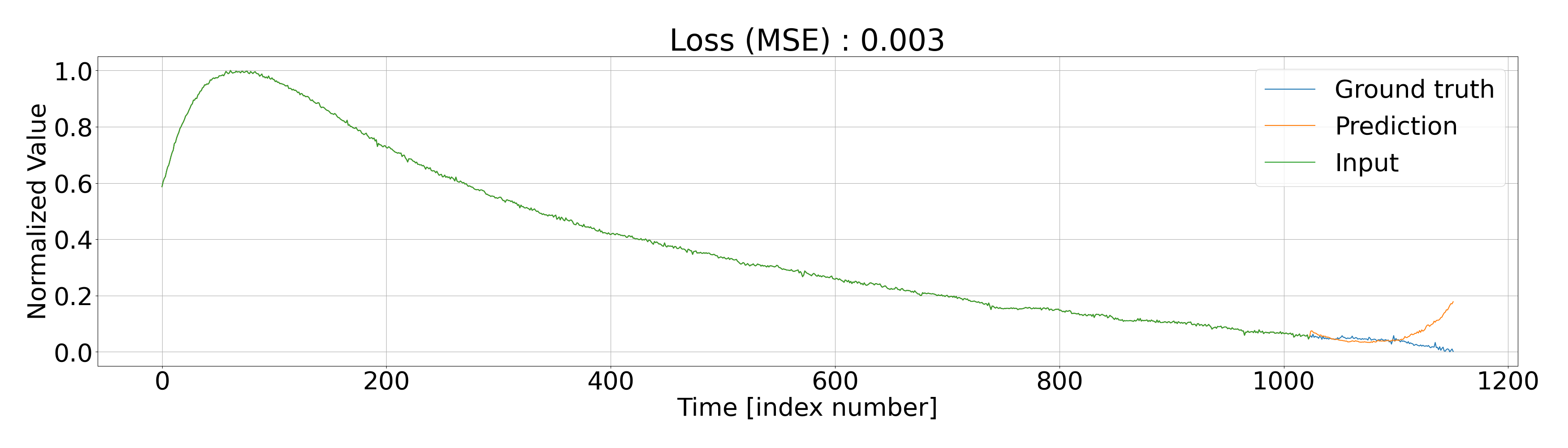}
        \caption{Corner case 3}
        \label{fig:subfig2}
    \end{subfigure}
    \caption{Examples of forecasts with corner cases. (a) Example from recording 1, (b) Example from recording 2, (c) Example from recording 3}
    \label{fig:mainfig}
\end{figure}

\subsection{Training datasets}
\label{sec:appendix_training}

\begin{figure}
    \centering
    \includegraphics[width=\linewidth]{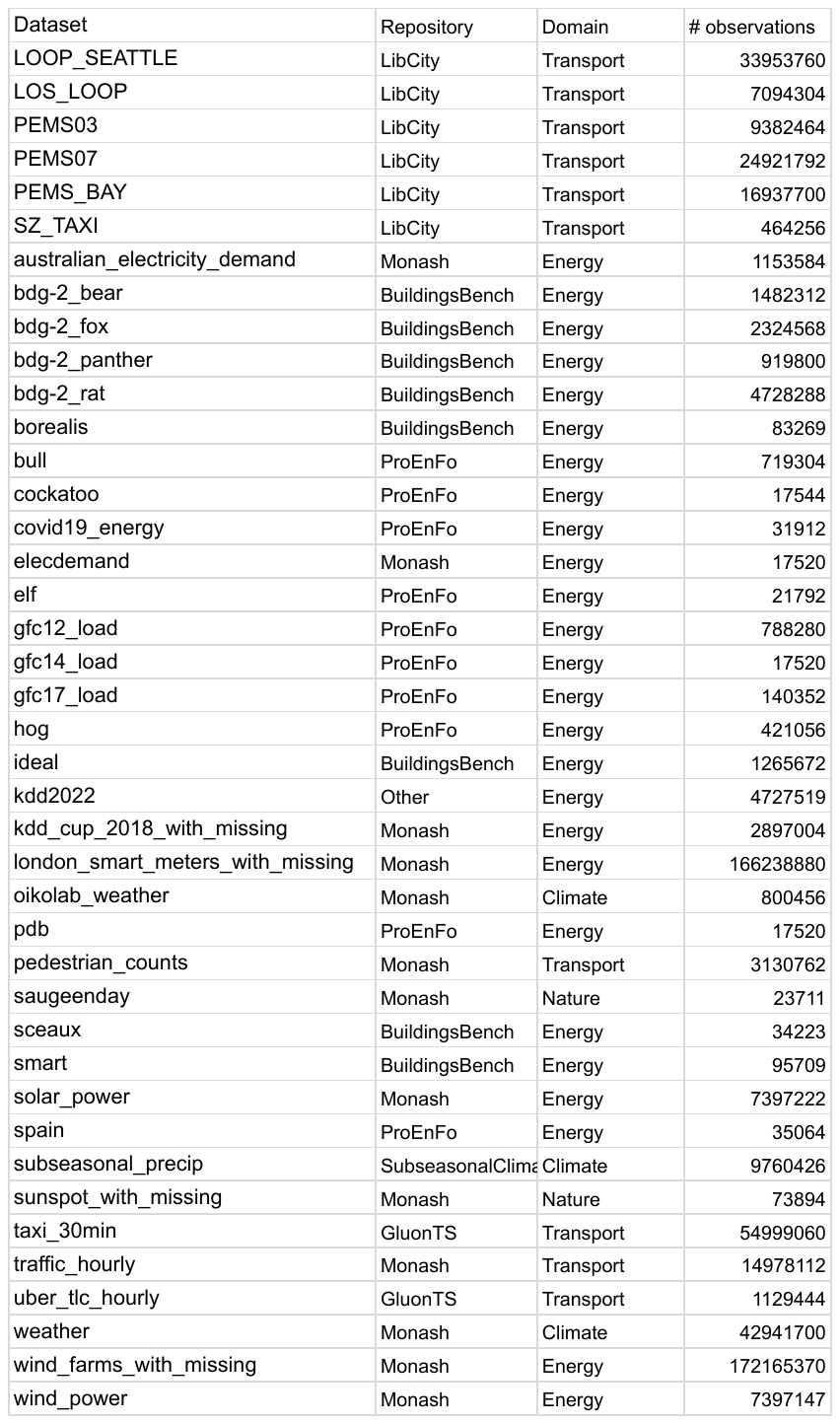}
    \caption{Datasets used for training}
    \label{fig:training_datasets}
\end{figure}

\end{document}